\documentclass[journal]{IEEEtran}

\usepackage{graphicx}
\usepackage{amsmath,amssymb,amsthm}
\usepackage{cite}
\usepackage{xcolor}

\usepackage{tikz}
\usetikzlibrary{arrows}
\usepackage{pgfplots}

\newtheorem{theorem}{Theorem}
\newtheorem{corollary}[theorem]{Corollary}

\newtheorem{proposition}[theorem]{Proposition}
\newtheorem{remark}[theorem]{Remark}

\newcommand{\norm}[1]{\Vert#1\Vert}

\usepackage[mathscr]{euscript}

\newcommand{\cR}{\mathcal{R}}

\newcommand{\R}{\mathbb{R}}

\newcommand{\E}{\mathbb{E}}
\renewcommand{\d}{\mathrm{d}}

\newcommand{\SIP}{{\rm SIP}}
\newcommand{\GIP}{\mathrm{GIP}_{\mathrm{1D}}}
\newcommand{\EGIP}{\mathrm{GIP}_{\mathrm{2D}}}

\newcommand{\NMI}{\mathrm{NMI}}
\newcommand{\ENMI}{\mathrm{ENMI}_{\mathrm{1D}}}
\newcommand{\DENMI}{\mathrm{ENMI}_{\mathrm{2D}}}

\renewcommand{\H}{{\rm H}}

\graphicspath{{FiguresJournal/}}

\title{Camera-Based Localization and\\Enhanced Normalized Mutual Information}

\author{\IEEEauthorblockN{Vishnu Teja Kunde,
Jean-Francois Chamberland,
Siddharth Agarwal}
\thanks{
This material is based upon work supported, in part, by Ford Autonomous Vehicles.
The first two authors are with the Department of Electrical and Computer Engineering, Texas A\&M University, College Station, TX, USA (emails: \{vishnukunde, chmbrlnd\}@tamu.edu).
Siddharth Agarwal is with Ford Motor Company, Dearborn, MI, USA (email: sagarw20@ford.com).
Portions of this research were conducted with the advanced computing resources provided by Texas A\&M High Performance Research Computing.}}

\pgfplotsset{compat=1.15} 

\begin{document}

\maketitle

\begin{abstract}
Robust and fine localization algorithms are crucial for autonomous driving.
For the production of such vehicles as a commodity, affordable sensing solutions and reliable localization algorithms must be designed.
This work considers scenarios where the sensor data comes from images captured by an inexpensive camera mounted on the vehicle and where the vehicle contains a fine global map.
Such localization algorithms typically involve finding the section in the global map that best matches the captured image.
In harsh environments, both the global map and the captured image can be noisy.
Because of physical constraints on camera placement, the image captured by the camera can be viewed as a noisy perspective transformed version of the road in the global map.
Thus, an optimal algorithm should take into account the unequal noise power in various regions of the captured image, and the intrinsic uncertainty in the global map due to environmental variations.
This article briefly reviews two matching methods: (i) standard inner product (SIP) and (ii) normalized mutual information (NMI).
It then proposes novel and principled modifications to improve the performance of these algorithms significantly in noisy environments.
These enhancements are inspired by the physical constraints associated with autonomous vehicles.
They are grounded in statistical signal processing and, in some context, are provably better.
Numerical simulations demonstrate the effectiveness of such modifications.
\end{abstract}

\section{Introduction}

Autonomous driving often relies on a structured hierarchical framework that may include sensing, localization, planning, control, and management \cite{mac2016heuristic,kuutti2018survey,yurtsever2020survey,kuutti2020survey,sun20214d,betz2022autonomous}.
Each of these components is critical in enabling vehicular autonomy and, hence, has received much attention in the literature.
There are many different sensing modalities and approaches that can inform localization in autonomous systems, ranging from global navigation satellite system (GNSS) and real-time kinematic (RTK) data to inertial measurement unit (IMU), standard camera images, LiDAR point clouds assisted with high-definition maps, and wireless signals \cite{schwarz2010mapping,joubert2020developments,humphreys2020deep,you2020Lidar,renzler2020increased,li2020lidar,sun2020mimo,you2023emergent}.
Moreover, it is worth noting that the perception system may be entrusted with several concurrent tasks such as the detection of surrounding vehicles, the evaluation of road conditions, and obstacle avoidance.
Having a multi-objective challenge can influence the design of the vehicle itself and, therefore, the character of the sensor observations.

This area has fueled many research initiatives in robotics and, more recently, vehicular autonomy.
Indeed, autonomous vehicles have benefited from  advances in the field of perception, decision-making, and planning.
Comprehensive surveys on the state-of-the-art and open challenges abound, e.g.~\cite{schwarting2018planning,janai2020computer,svarm2016city}.
Furthermore, localization is closely related to simultaneous localization and mapping (SLAM) in terms of technology \cite{durrant2006simultaneous,bailey2006simultaneous,ferris2007wifi}, although the problem at hand is conceptually simpler.
One major distinction between SLAM and localization, as treated in this article, is the availability of highly detailed maps at the autonomous vehicle~\cite{reid2019localization};
whereas, in SLAM, a global map is built incrementally as the vehicle navigates to unknown areas~\cite{dissanayake2001solution,eliazar2003dp,bosse2004simultaneous,thrun2006graph,klein2007parallel,engel2014lsd,mur2015orb}.
This latter body of academic work is nevertheless pertinent because the estimation portion of a SLAM algorithm can provide insight into efficient fusion methods for standard localization.

For most self-driving applications, stand-alone positioning systems such as global navigation satellite system (GNSS) and real-time kinematic positioning (RTK) are inadequate, only offering coarse localization~\cite{levinson2010robust,levinson2011towards,bresson2017simultaneous}.
In high traffic scenario such as urban environments, autonomy often requires vehicles to have (at least) centimeter-scale positional accuracy.
Hence, fine localization involving sensory data as input is crucial, as it offers additional information to the vehicle.
Specifically, image matching often becomes a key component of a complex localization system, with additional modules such as an extended Kalman filter (EKF) based on RTK, IMU data and a wireless ultra-wideband radar.
Motivated by the quest for greater accuracy, we revisit the structure of established approaches and we identify algorithmic improvements that boost performance.
While image matching may only be one of several components in an typical localization scheme, our work focuses exclusively on this important building block.
For ease of exposition, we center our discussion on this single aspect, with the understanding that improvements in image matching will eventually lead to better accuracy in more complex implementations as well.
Throughout, we assume that an accurate global map, perhaps in the form of a high-resolution infrared remittance ground map, is available at the vehicle.

A conceptual starting point for this exposition is an image-based localization scheme that calculates the standard Euclidean distance between the camera image and a candidate location on the global map~\cite{kruger1998image,konolige1999markov}.
Under certain assumptions, this process yields the likelihood of the vehicle being at a particular location.
This action is repeated at several possible locations to then estimate the true position of the vehicle.
During this process, images captured by vehicle-mounted cameras are rectified to accommodate the disparate viewpoints between the vehicle and the bird's-eye view of the global map~\cite{furgale2013toward,heng2019project}.
This action results in uneven noise amplification across pixels, quantitatively captured by the signal-to-noise ratio (SNR), and consequently invites the application of advanced statistical signal processing techniques, beyond the standard inner product.
Unfortunately, these technical subtleties are not preserved by computer vision libraries and tools when applying perspective transformations.
As we will see, this situation warrants careful attention and the eventual application of a weighted inner product when computing image distances.
Harnessing noise properties and characterizing signal structures is a topic of broad interest in the signal processing community, with multiple significant contributions, e.g., see~\cite{wax1985detection,feder1988parameter,rao1992model}.
Well-studied examples of this phenomenon include source localization, imaging, and tracking~\cite{ding1987generalized,manolakis2002detection,taj2011distributed,ravishankar2012learning}.
Techniques akin to those leveraged in this article have been employed to remove colored noise from audio signals~\cite{boll1979suppression,scalart1996speech,kamath2002multi} and non-uniform noise from images~\cite{chehdi1992new,hirakawa2006joint,ghazal2010homogeneity,rajwade2012image}.
Such approaches share many notional similarities with our proposed scheme, albeit mostly at a higher conceptual level.

Building on the idea of statistical confidence across regions of the signal, we look into the possible improvement of another existing similarity measure for signal matching known as normalized mutual information (NMI), which has been proposed for image registration in medical diagnosis~\cite{pluim2000image,maes2003medical} and has since been applied in broader contexts.
An important feature of NMI matching in image classification is its robustness to intensity shifts between two images.
However, current implementations of NMI for autonomous localization also seem to disregard the uneven noise distribution across the regions of the image, giving absolute confidence to the acquired image, equally across the regions.
As mentioned above, this uneven noise distribution is inherent to the problem because of the presence of a shifted perspective of the mounted camera in obtaining the features that describe the scene used for localization.

The geometric properties of mounted cameras and their impact on pixel quality create an opportunity for algorithmic improvement in the following sense.
With the same data acquisition device, vehicle, and global map; it is often possible to improve the performance of existing algorithms by incorporating notions from statistical signal processing.
The improvements can potentially be implemented in software and deployed as a system update.
Below, we discuss the mathematical underpinnings more precisely, we described the enhanced algorithms, and we report offer empirical evidence on potential performance improvements.
The statistical signal processing concepts behind the scenarios considered below are likely to find broader applicability in autonomous vehicles; there are many localization algorithms, yet the physical properties of image acquisition by vehicles are governed by the same constraints.

\section{Naive Model and Problem Formulation}

We examine vision-based localization of a ground vehicle for which the sensory data is the image acquired using a forward-facing camera (see Fig.~\ref{fig:RoadCoorSys}).
The algorithm seeks to match the captured image with sections of the global map.
The data corresponding to the section having the best match with the captured image is employed to estimate the location of the vehicle.
The captured image needs to be appropriately processed in order to be matched with the global map.
This requires, at its core, a certain perspective transformation whose geometry we describe in the next section.

\subsection{Focal Plane Geometry}

Consider the pinhole model of the camera mounted on the vehicle, with its 2D focal plane represented by the coordinates $(\tilde{x}, \tilde{y})$.
A location in the space in front of the vehicle can be specified using the 3D ``pinole'' coordinate system $(x, y, z)$, as illustrated in Fig.~\ref{fig:FocalPlaneSys}.
Let $f$ be the focal length of the camera.
Then, in the absence of occlusions, coordinate $(x,y,z)$ will project onto the focal point $(\tilde{x}, \tilde{y})$ according to the transformation
\begin{xalignat}{2} \label{eq:focalplane}
    \tilde{x} &= \frac{f x}{z} &
    \tilde{y} &= \frac{f y}{z}.
\end{xalignat}
\begin{figure}
    \centering{\begin{tikzpicture}[
  font=\footnotesize, >=stealth', line width=1.0pt, line cap=round
]

\def\mytheta{-35}

\draw (-3, 0.9) node[anchor=south] {2D Focal Plane};
\draw (-3, 0.5) node[anchor=south] {Coordinate System};
\draw (-3.5, -0.5) rectangle (-2.5, 0.5);
\draw[->] (-3, 0) -- (-3, -1) node[anchor=east] {$\tilde{y}$};
\draw[->] (-3, 0) -- (-2, 0) node[anchor=north] {$\tilde{x}$};

\draw[rotate around={\mytheta:(0, 0)}] (-1, -0.5) rectangle (0, 0.5);
\draw[rotate around={\mytheta:(0, 0)}] (-1, 0) -- (1.25,0);
\draw[gray, dashed, rotate around={\mytheta:(0, 0)}] (-1, 0.2) -- (1.5,-0.3);

\draw[draw=gray,fill=gray,rotate around={\mytheta:(0, 0)}] (1.5,-0.3) circle [radius=1.25pt];
\draw[draw=gray,fill=gray,rotate around={\mytheta:(0, 0)}] (-1, 0.2) circle [radius=0.75pt];
\node at (1.75, -1.1) {$(x,y,z)$};

\node (pinhole) at (0, -1) {Pinhole}
  edge[->] (0,-0.1);
\draw[rotate around={\mytheta:(0, 0)}] (-1, 0) node[anchor=south, rotate=90+\mytheta] {Focal Plane};


\draw (3, 0.9) node[anchor=south] {3D Pinhole};
\draw (3, 0.5) node[anchor=south] {Coordinate System};
\draw (3.5, -0.5) circle (0.15) node[anchor=north west] {$z$};
\draw (3.65, -0.5) -- (3.35, -0.5);
\draw (3.5, -0.65) -- (3.5, -0.35);
\draw[->] (3.5, -0.3) -- (3.5, 0.5) node[anchor=north west] {$y$};
\draw[->] (3.5, -0.5) -- (2.5, -0.5) node[anchor=south west] {$x$};

\end{tikzpicture}}
    \caption{This notional diagram shows the (standard) alignment between the external frame of reference of the camera (right) whose origin is at the pinhole, and the internal focal plane of the camera (left).}
    \label{fig:FocalPlaneSys}
\end{figure}
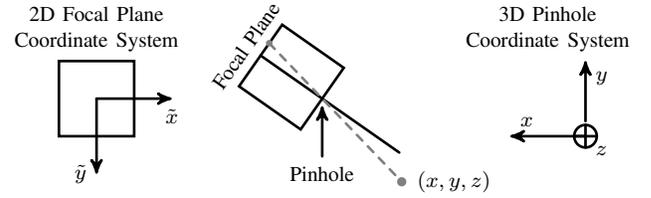
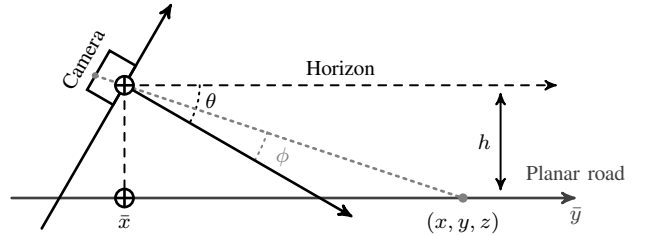
\begin{figure}
\centering{\begin{tikzpicture}[
  font=\footnotesize, >=stealth', line width=1.0pt, line cap=round
]

\def\mytheta{-30}

\draw[->, darkgray] (-1.5, -1.5) -- (6, -1.5) node[anchor=north] {$\bar{y}$}
node[anchor=west, above, yshift=1mm] {Planar road} ;

\draw[<->, line width=0.75] (5, -0.1) -- (5, -1.4) node[midway,left] {$h$};

\draw[dotted, line width=0.75] (1, 0) arc (0:\mytheta:1);
\node[text=black] at (1.15, -0.2) {$\theta$} ;

\draw[dashed, line width=0.75] (0, 0) -- (0, -1.5);
\draw[fill=white] (0, -1.5) circle (1.2mm) node[anchor=north, yshift=-1mm] {$\bar{x}$};
\draw (-0.12,-1.5) -- (0.12,-1.5);
\draw (0, -1.5-0.12) -- (0, -1.5+0.12);
\node at (4.5, -1.825) {$(x,y,z)$};

\draw[dashed, ->, line width=0.75] (0, 0) -- (5.75, 0) node[midway, above] {Horizon} ;

\draw[rotate around={\mytheta:(0, 0)}] (-0.4, -0.3) rectangle (0, 0.3);
\node[rotate={90+\mytheta}] at (-0.55, 0.35) {Camera};


\draw[gray,dotted,-] (-0.39, 0.13) -- (4.5,-1.5);
\draw[draw=gray,fill=gray] (4.5,-1.5) circle [radius=1.25pt];
\draw[draw=gray,fill=gray] (-0.39, 0.13) circle [radius=0.75pt];
\draw[dotted, gray, line width=0.75] (1.732, -1) arc (\mytheta:-17.43:2);
\node[text=black, gray] at (2.1,-0.95) {$\phi$} ;

\draw[->] (0, 0) -- (\mytheta+90:1.25); 
\draw[-] (0, 0) -- (\mytheta-90:2.2);
\draw[->] (0, 0) -- (\mytheta:3.5); 
\draw[fill=white] (0,0) circle (1.2mm); 
\draw (-0.12,0) -- (0.12,0);
\draw (0,-0.12) -- (0,0.12);

\end{tikzpicture}}
    \caption{The above figure illustrates the coordinate system of the view in front of the camera $(x, y, z)$ and the (planar) 2D coordinate system of the road $(\bar{x}, \bar{y})$.}
    \label{fig:RoadCoorSys}
\end{figure}

For the purpose of localization, we are interested in coordinates that are confined to the road surface, which is assumed to form a horizontal plane.
Let $(\bar{x}, \bar{y})$ represent the coordinates of a point on the road surface, as depicted in Fig.~\ref{fig:RoadCoorSys}.
Assume $h$ is the height at which the camera is mounted above the surface of the road, and let $\theta$ be the angle of depression of the camera from the the horizontal line parallel to the surface of the road.
Since the road represents a plane in the 3D space, the surface only has two degrees of freedom: $\bar{x}$ and $\bar{y}$ or, equivalently, $\bar{x}$ and $\phi$.
We wish to find how a point in the $\bar{x}\bar{y}$-plane maps to the coordinate system of the view in front of the camera.

\begin{figure}
\centering{\begin{tikzpicture}[
  font=\footnotesize, >=stealth', line width=1.0pt, line cap=round
]

\def\mytheta{-30}

\draw[dotted, line width=0.75, fill=lightgray] (0,0) -- (1, 0) arc (0:\mytheta:1) -- (0,0);
\node[text=black] at (1.15, -0.2) {$\theta$} ;

\draw[line width=0.75] (0,0) -- node[midway,above]{$\bar{y}$} (4.5,0) -- node[midway,right] {$h$} (4.5,-1.5) -- (0,0);
\draw[|-, line width=0.5] (0,0.25) -- (2, 0.25);
\draw[-|, line width=0.5] (2.5,0.25) -- (4.5, 0.25);
\draw[|-, line width=0.5] (4.75,0) -- (4.75, -0.5);
\draw[-|, line width=0.5] (4.75,-1) -- (4.75, -1.5);


\draw[line width=1.0] (0,0) -- (4.5,-1.5);

\draw[dashed,-] (4.5,-1.5) -- node[midway,left] {$y$} (4.09332,-2.32) -- node[midway,below] {$z$}(0,0);
\draw[draw=black,fill=black] (4.5,-1.5) circle [radius=1.25pt];
\draw[draw=black,fill=black] (0, 0) circle [radius=1.25pt];
\draw[dotted, line width=0.75] (1.732, -1) arc (\mytheta:-17.43:2);
\node at (2.1,-0.95) {$\phi$} ;

\draw[dotted] (4.5,-1.5) -- (4.5,-2.32) -- (0,-2.32) -- (0,0);

\end{tikzpicture}}
    \caption{The fundamental triangles that govern the projection of points on the road to the focal plane of the camera. The two right triangles share a hypotenuse, which can be leveraged to compute pertinent quantities.}
    \label{fig:FundamentalTraiangles}
\end{figure}
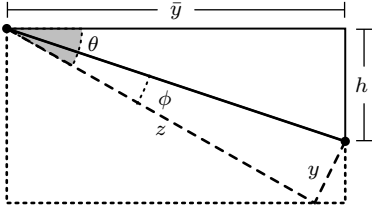
By construction, the first relation is simply $\bar{x} = x$.
To derive the second governing equation, it may be useful to refer to Fig.~\ref{fig:FundamentalTraiangles}, which isolates the fundamental triangles that connect $\bar{y}$, $y$, and $z$.
Therein, $\theta$ corresponds to the shaded angle.
The length of the shared hypotenuse is 
\begin{equation}
\sqrt{y^2 + z^2} = \sqrt{h^2 + \bar{y}^2} .
\end{equation}
Looking at the top triangle, we have $\tan (\theta - \phi) = {h}/{\bar{y}}$, which yields an explicit expression for $\phi$,
\begin{equation}
\phi = \theta - \arctan \left( \frac{h}{\bar{y}} \right) .
\end{equation}
The ratio $y/z$, which appears in \eqref{eq:focalplane}, can be obtained using the general trigonometric identity below,
\begin{equation}
\tan (\theta - \psi) = \frac{\tan \theta - \tan \psi}{1 + \tan \theta \tan \psi} .
\end{equation}
Applying this result produces the ratio
\begin{equation}
\begin{split}
\frac{y}{z} &= \tan \phi = \tan \left( \theta - \arctan \left( \frac{h}{\bar{y}} \right) \right) \\
&= \frac{\tan \theta - \frac{h}{\bar{y}}}{1 + \frac{h}{\bar{y}} \tan \theta}
= \frac{\bar{y} \tan \theta - h}{\bar{y} + h \tan \theta} .
\end{split}
\end{equation}
Likewise, we can leverage the general trigonometric identity
\begin{equation}
\cos (\theta - \psi) = \cos \theta \cos \psi + \sin \theta \sin \psi
\end{equation}
and write
\begin{equation}
\begin{split}
\frac{x}{z} &= \frac{\bar{x}}{\sqrt{h^2 + \bar{y}^2} \cos \phi} \\
&= \frac{\bar{x}}{\sqrt{h^2 + \bar{y}^2} \cos \left( \theta - \arctan \left( \frac{h}{\bar{y}} \right) \right)} \\
&= \frac{\bar{x}}{\bar{y} \cos \theta + h \sin \theta} .
\end{split}
\end{equation}
Thence, we obtain
\begin{align}
\tilde{x} &= \frac{f x}{z}
= \frac{f \bar{x}}{\bar{y} \cos \theta + h \sin \theta} \\
\tilde{y} &= \frac{f y}{z}  
= f \frac{\bar{y} \tan \theta - h}{\bar{y} + h \tan \theta}
= f \frac{\bar{y} \sin \theta - h \cos \theta}{\bar{y} \cos \theta + h \sin \theta} .
\end{align}

Having established a mathematical relation between points on the road and their projection on the focal plane of the camera, we turn to the characterization of noise.
To do so, we must compute the Jacobian determinant of the transformation.
This can be accomplished by calculating the partial derivatives, first with respect to $\bar{x}$,
\begin{xalignat}{2}
\frac{\partial \tilde{x}}{\partial \bar{x}} &= \frac{f}{\bar{y} \cos \theta + h \sin \theta} &
\frac{\partial \tilde{y}}{\partial \bar{x}} &= 0
\end{xalignat}
and also with respect to $\bar{y}$,
\begin{align}
\frac{\partial \tilde{x}}{\partial \bar{y}} &= - \frac{f \bar{x} \cos \theta}{\left( \bar{y} \cos \theta + h \sin \theta \right)^2} \\
\begin{split}
\frac{\partial \tilde{y}}{\partial \bar{y}}
&= f \frac{\sin \theta}{\bar{y} \cos \theta + h \sin \theta}
- f \cos \theta \frac{\bar{y} \sin \theta - h \cos \theta}{\left( \bar{y} \cos \theta + h \sin \theta \right)^2} \\
&= \frac{f h}{\left( \bar{y} \cos \theta + h \sin \theta \right)^2} .
\end{split}
\end{align}
The corresponding Jacobian matrix is given by
\begin{equation*}
\begin{split}
\mathbf{J} &= \begin{bmatrix} \frac{\partial \tilde{x}}{\partial \bar{x}} & \frac{\partial \tilde{y}}{\partial \bar{x}} \\
\frac{\partial \tilde{x}}{\partial \bar{y}} & \frac{\partial \tilde{y}}{\partial \bar{y}} \end{bmatrix}
= \begin{bmatrix} \frac{f}{\bar{y} \cos \theta + h \sin \theta} & 0 \\
- \frac{f \bar{x} \cos \theta}{\left( \bar{y} \cos \theta + h \sin \theta \right)^2}
& \frac{f h}{\left( \bar{y} \cos \theta + h \sin \theta \right)^2} \end{bmatrix}
\end{split}
\end{equation*}
and we can compute its determinant as
\begin{equation} \label{eq:detJ}
\mathrm{det}(\mathbf{J}) = \frac{f^2 h}{\left( \bar{y} \cos \theta + h \sin \theta \right)^3}.
\end{equation}

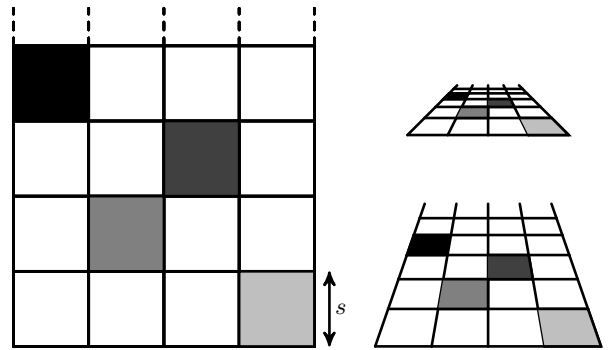
\begin{figure}[h!]
\centering{
{{\begin{tikzpicture}[
  font=\small, >=stealth', line width=1.0pt, line cap=round, scale=1
]

\foreach \z in {1,2,3,4,5} {
  \foreach \x in {-2,-1,0,1,2} {
    \draw (\x, 1) -- (\x, 5);
    \draw[dashed] (\x, 5) -- (\x, 5.5);
    \draw (-2, \z) -- (2, \z);
  }
}
\draw[<->] (2.2, 1) to (2.2, 2);
\node at (2.35, 1.5){$s$};

\draw[fill=lightgray] (1,1) rectangle (2,2);
\draw[fill=gray] (-1,2) rectangle (0,3);
\draw[fill=darkgray] (0,3) rectangle (1,4);
\draw[fill=black] (-2,4) rectangle (-1,5);

\end{tikzpicture}}} 
{{\begin{tikzpicture}[
  font=\small, >=stealth', line width=1.0pt, line cap=round, scale=1
]



\foreach \mytheta/\d in {60/3.5,80/0} {
    \foreach \y in {2,3,4,5,6,7} {
      \foreach \x in {-2,-1,0,1} {
        \draw ({\x / (\y * cos(\mytheta) + sin(\mytheta))}, {(\y * tan(\mytheta) - 1) / (\y + tan(\mytheta)) + \d})
        -- ({(\x+1) / (\y * cos(\mytheta) + sin(\mytheta))}, {(\y * tan(\mytheta) - 1) / (\y + tan(\mytheta)) + \d});
      }
    }
    \foreach \y in {2,3,4,5,6,7} {
      \foreach \x in {-2,-1,0,1,2} {
        \draw ({\x / (\y * cos(\mytheta) + sin(\mytheta))}, {(\y * tan(\mytheta) - 1) / (\y + tan(\mytheta)) + \d})
        -- ({\x / ((\y+1) * cos(\mytheta) + sin(\mytheta))}, {((\y+1) * tan(\mytheta) - 1) / ((\y+1) + tan(\mytheta)) + \d});
      }
    }
    
    \draw[fill=black] ({-1 / (5 * cos(\mytheta) + sin(\mytheta))}, {(5 * tan(\mytheta) - 1) / (5 + tan(\mytheta)) + \d})
    -- ({-2 / (5 * cos(\mytheta) + sin(\mytheta))}, {(5 * tan(\mytheta) - 1) / (5 + tan(\mytheta)) + \d})
    -- ({-2 / (6 * cos(\mytheta) + sin(\mytheta))}, {(6 * tan(\mytheta) - 1) / (6 + tan(\mytheta)) + \d})
    -- ({-1 / (6 * cos(\mytheta) + sin(\mytheta))}, {(6 * tan(\mytheta) - 1) / (6 + tan(\mytheta)) + \d});
    
    \draw[fill=darkgray] ({0 / (4 * cos(\mytheta) + sin(\mytheta))}, {(4 * tan(\mytheta) - 1) / (4 + tan(\mytheta)) + \d})
    -- ({1 / (4 * cos(\mytheta) + sin(\mytheta))}, {(4 * tan(\mytheta) - 1) / (4 + tan(\mytheta)) + \d})
    -- ({1 / (5 * cos(\mytheta) + sin(\mytheta))}, {(5 * tan(\mytheta) - 1) / (5 + tan(\mytheta)) + \d})
    -- ({0 / (5 * cos(\mytheta) + sin(\mytheta))}, {(5 * tan(\mytheta) - 1) / (5 + tan(\mytheta)) + \d});
    
    \draw[fill=gray] ({-1 / (3 * cos(\mytheta) + sin(\mytheta))}, {(3 * tan(\mytheta) - 1) / (3 + tan(\mytheta)) + \d})
    -- ({0 / (3 * cos(\mytheta) + sin(\mytheta))}, {(3 * tan(\mytheta) - 1) / (3 + tan(\mytheta)) + \d})
    -- ({0 / (4 * cos(\mytheta) + sin(\mytheta))}, {(4 * tan(\mytheta) - 1) / (4 + tan(\mytheta)) + \d})
    -- ({-1 / (4 * cos(\mytheta) + sin(\mytheta))}, {(4 * tan(\mytheta) - 1) / (4 + tan(\mytheta)) + \d});
    
    \draw[fill=lightgray] ({1 / (2 * cos(\mytheta) + sin(\mytheta))}, {(2 * tan(\mytheta) - 1) / (2 + tan(\mytheta)) + \d})
    -- ({2 / (2 * cos(\mytheta) + sin(\mytheta))}, {(2 * tan(\mytheta) - 1) / (2 + tan(\mytheta)) + \d})
    -- ({2 / (3 * cos(\mytheta) + sin(\mytheta))}, {(3 * tan(\mytheta) - 1) / (3 + tan(\mytheta)) + \d})
    -- ({1 / (3 * cos(\mytheta) + sin(\mytheta))}, {(3 * tan(\mytheta) - 1) / (3 + tan(\mytheta)) + \d});
}

\end{tikzpicture}}}}
\caption{The left diagram shows an arbitrary grid of equally sized squares on the road in front of the autonomous vehicle.
The right diagrams show the grid pattern's image on the focal plane of the camera.
Squares with the same physical area do not project onto equally sized sections of the focal plane.
The two diagrams on the right are for different values of $\theta$.
}
\label{fig:Grid2}
\end{figure}

We illustrate how this transformation affects captured areas with the following example.
Let $\bar{\mathcal{R}} = [\bar{x}_\ell, \bar{x}_u]\times[\bar{y}_\ell, \bar{y}_u] \in \R^2$ be a rectangular region on the surface of the road.
We refer to such regions as \emph{tiles} because they form the elements of a tessellation of the road.
Let $\tilde{\mathcal{R}}$ be the representation of the region on the focal plane of the camera. Let $\bar{{A}}, \tilde{{A}}$ denote the areas of $\bar{\mathcal{R}}$ and $\tilde{\mathcal{R}}$, respectively.
Then, we have
\begin{align} \label{eq:FocalPlaneArea}
\begin{split}
&\tilde{{A}} = \underset{\tilde{\mathcal{R}}}{\iint} \d\tilde{x}\d\tilde{y}
= \underset{\bar{\mathcal{R}}}{\iint} \vert\mathrm{det}(\mathbf{J})\vert \d\bar{x} \d\bar{y} \\
&= \int_{\bar{x}_\ell}^{\bar{x}_u} \d \bar{x} \int_{\bar{y}_\ell}^{\bar{y}_u} \frac{f^2 h}{\left( \bar{y} \cos \theta + h \sin \theta \right)^3} \d\bar{y} \\
&= \frac{\bar{x}_u - \bar{x}_\ell}{2 \cos \theta}
\left[ \frac{f^2 h}{\left( \bar{y}_\ell \cos \theta + h \sin \theta \right)^2} - \frac{f^2 h}{\left( \bar{y}_u \cos \theta + h \sin \theta \right)^2} \right] .
\end{split}
\end{align}
That is, a rectangular region of area $\bar{{A}} = (\bar{x}_u-\bar{x}_\ell)(\bar{y}_{u}-\bar{y}_\ell)$ on the road ahead gets transformed into a distorted region on the focal plane of the camera.
Observe that the area decreases rapidly as a function of the distance $\bar{y}$.
This is illustrated in Fig.~\ref{fig:Grid2} where a road is depicted along with its projection onto the focal plane of a camera for two distinct camera angles.

\section{Image Acquisition and Noise Model}
\label{section:ImageAcquisition}

We now describe the noisy image acquisition process that affects the quality of the information across the regions of the captured image.
Assume that, at a given time and location, the features on the road surface on the global map at point $(\bar{x}, \bar{y})$ is given by the intensity function $g(\bar{x}, \bar{y})$.
This is the signal that captures the \emph{variations} of the intensity across the road surface (and so is assumed to be zero-mean).
Consider a situation where the signal is constant over rectangular regions or tiles, as described in the previous section.
Furthermore, assume that the signal values are independent across \emph{tiles}.
In this illustrative example, we take the road surface to be tessellated into square tiles of side $s$ (see Fig.~\ref{fig:Grid2}).
Let $w$ be the width of the road, and let $d$ be the length of the road section in the line of sight.
In this didactic setting, there are $N_w \times N_d$ tiles in the image captured by the mounted camera. Denote $N_w = \lceil \frac{w}{s} \rceil, N_d = \lceil\frac{d}{s} \rceil$ to be the number of tiles across and along the road respectively as per the above tessellation.
Therefore, this tiled road surface can be captured using the equation
\begin{equation*}
g(\bar{x}, \bar{y}) = \sum_{k=1}^{N_w} \sum_{j=1}^{N_d}(a_{k,j} + n_{k,j}^i) \mathbf{1}_{\{(\bar{x}, \bar{y}) \in  \bar{\mathcal{R}}_{k,j}\}},
\end{equation*}
where road region $\bar{\mathcal{R}}_{k,j} \triangleq [(k-1)s, ks] \times [(j-1)s, js]$, for $k \in [N_w]$ and $j \in [N_d]$.
Value $a_{k,j}$ can be viewed as a (fixed) realization of the long-term road surfacing process, whereas $n_{k,j}^i$ is the intrinsic noise that captures the short-term variation due to environmental conditions such as current illumination and the presence of dirt.
The purpose of the first component $\{ a_{k,j} \}$ is to capture the fundamental variations in a scene, which can depend on buildings, natural features, and road materials.
Every such variable is essentially constant over time, as it pertains to localization.
The second component $n_{k,j}^i$ accounts for rapidly changing visual attributes that can be affected by phenomena such as rain, wind, light, etc.
Specifically, $n_{k,j}^i$ will be different if the camera snaps pictures at two different times, whereas $a_{k,j}$ will remain essentially constant over the same period.
Let $\sigma^2 = \E \big[a_{k,j}^2 \big]$, and $\sigma_i^2 = \E \big[ \big( n_{k,j}^i \big)^2 \big]$ be the power of the signal and the intrinsic noise, respectively.
We define the signal to intrinsic noise ratio $(\mathrm{SINR})$ by
\begin{equation*}
    {\mathrm{SINR}} = \frac{\sigma^2}{\sigma_i^2}.
\end{equation*}



Turning to the camera's perspective, we assume that the charge-coupled device (CCD) of the camera is subject to an additive white Gaussian noise process $\tilde{N}(\tilde{x}, \tilde{y})$ over the captured section of the image, with power spectral density $N_0$.
The signal sensed on the focal plane of the camera within region $\tilde{\mathcal{R}}_{k,j}$ is given by
\begin{equation*}
    S_{\mathrm{ccd}}(k,j) =  \iint_{\tilde{\mathcal{R}}_{k,j}}  \tilde{g}(\tilde{x}, \tilde{y}) \d\tilde{x}\tilde{y} + \iint_{\tilde{\mathcal{R}}_{k,j}} \tilde{N}(\tilde{x}, \tilde{y}) \d \tilde{x} \d \tilde{y}, 
\end{equation*}
and so we can equivalently write
\begin{equation} \label{eq:Sccd}
S_{\mathrm{ccd}}(k,j) = a_{k,j} \tilde{A}_{k,j} + n_{k,j} \tilde{A}_{k,j} + n_s(\tilde{\cR}_{k,j})
\end{equation}
where the variance of $n_s(\tilde{\cR}_{k,j})$, the noise introduced by the CCD within the corresponding tile area, is equal to $N_0 \tilde{A}_{k,j}$.
The overall power in the first component of \eqref{eq:Sccd} is
\begin{equation}
\E[a_{k,j}^2] \tilde{A}_{k,j}^2 = \sigma^2 \tilde{A}_{k,j}^2 . 
\end{equation}
Hence, the signal to sensor noise ratio (${\rm SSNR}$) in the $(k,j)$th tile is given by
\begin{equation*}
\mathrm{SSNR}_{k,j} = \frac{\sigma^2 \tilde{A}_{k,j}}{N_0},
\end{equation*}
and the ${\rm SINR}$ in the same region is given by
\begin{equation*}
\mathrm{SINR}_{k,j} = \frac{\sigma^2}{\sigma_i^2}.
\end{equation*}

We emphasize that, according to \eqref{eq:FocalPlaneArea},
\begin{equation}
\begin{split}
\mathrm{SSNR}_{k,j}
= \frac{\sigma^2 f^2 h}{2 N_0 \cos \theta}
&\left[ \frac{s}{\left( (j-1)s \cos \theta + h \sin \theta \right)^2} \right. \\
&\qquad - \left. \frac{s}{\left( js \cos \theta + h \sin \theta \right)^2} \right] .
\end{split}
\end{equation}
Thus, the signal to sensor noise ratio $\mathrm{SSNR}_{k,j}$ decreases as the function of $j$.
That is, the acquired value of a tile that is located near the front of the car is much more reliable than the acquired value of a tile located far ahead.
This matches human intuition because farther tiles map to smaller areas, as seen on Fig.~\ref{fig:Grid2}, and hence their sensed values are less reliable.
The height of the camera $h$ and its angle of depression $\theta$ also have significant impact on the signal to sensor noise ratio.
For the latter parameter, this can be seen on the right side of Fig.~\ref{fig:Grid2}.

A final component of image acquisition that does not appear in our discussion is the actual size of the image sensor on the focal plane of the camera.
The hardware physically limits the part of the scene that can be acquired.
This too will be severly affected by $h$ and $\theta$.
While this is an important engineering consideration, it does not enter in the signal processing techniques we wish to develop.
Rather, this reality comes as a design constraint that must be accounted for once the effect of noise amplification is understood.

Fundamentally, \eqref{eq:detJ} and \eqref{eq:FocalPlaneArea} describe mathematically how sections of a same image can have different reliability.
When an image is rectified through a perspective transformation, the squares in Fig.~\ref{fig:Grid2} can go back to looking like a grid; however, this invertible transformation does not alter the effective values of the signal to sensor noise ratio associated with various squares.
Indeed, the projection onto the focal plane of the camera followed by a perspective transformation leads to unequal noise amplification.
This, in turn, affects the performance of matching algorithms, as described in the next section.

\section{Detection and Improvements}

At an abstract level, the localization algorithms we wish to study involve matching a captured image with a section of the global map after a perspective transformation.
For simplicity, we assume that the global map is quantized into a set of $L$ possible sections of fine locations.
As described earlier, every section and the captured image have $N_w \times N_d$ tiles.
Therefore, we can view the tiles of the captured image as an $N_w\times N_d$ dimensional matrix, $\mathbf{Y} = \mathbf{A} + \mathbf{N}^i + \mathbf{N}^s$, where $a_{k, j}$ is the signal amplitude, $n^i_{k,j}$ is the intrinsic noise, $n^s_{k,j}$ is the sensor noise in the $(k,j)$th tile of the acquired image; the variances of $n^i_{k,j}$ and $n^s_{k,j}$ are given by $(\sigma_{k, j}^i)^2 = \sigma_i^2$ and $(\sigma^s_{k,j})^2 = N_0/\tilde{A}_{k,j}$, respectively.

Similarly, the $\ell$th section in the global map can be represented as $\mathbf{Y}^\ell = \mathbf{A}^\ell + \mathbf{N}^{i, \ell}$, where $n_{k, j}^{i, \ell}$ is the intrinsic noise that has zero mean and variance $(\sigma_{k, j}^i)^2 =\sigma_i^2$.
We assume that, if the true location of the captured image is $\ell$, then $\mathbf{A} = \mathbf{A}^{\ell}$.
For a matrix $\mathbf{Y}$, we denote its (column-wise) vectorized form as $\vec{\mathbf{y}}$, with
\begin{equation} \label{eq:vectorize}
\vec{y}_{k + (j-1) N_w} = y_{k,j} \quad k \in [N_w], j \in [N_d] .
\end{equation}

The tiling approach adopted herein is inspired by the quantization that takes place in certain algorithms employed in industry.
We begin our exposition with a didactic scenario: maximum likelihood detection.
This setting is simple, yet it captures the statistical properties we wish to leverage in our proposed algorithmic improvements.

\subsection{Maximum Likelihood Detection}
\label{subsection:MaximumLikelihoodDetection}


In this section, we examine the process of matching the captured image to a section of the global map using the Euclidean distance between the tiled images in the value space.
We declare the section with the smallest Euclidean distance as the match and use the metadata associated with this section to produce an estimate of the location of the vehicle.
Note that the Euclidean norm is the norm induced by the standard inner product (${\rm SIP}$) and, therefore, we can write the estimate of the location as
\begin{equation} \label{eq:EuclideanNorm}
\begin{split}
\hat{\ell}_{\SIP}
&= \operatorname*{arg\;min}_{\ell\in[L]} \norm{\mathbf{Y}-\mathbf{Y}^\ell}_{\mathrm{F}} \\
&= \operatorname*{arg\;min}_{\ell\in[L]} \norm{\vec{\mathbf{y}}-\vec{\mathbf{y}}^\ell}_2
\end{split}
\end{equation}
where the first equality is the matrix Frobenius norm, and the second equality is the standard Euclidean norm applied to the vectorized form.
The above detection algorithm implicitly gives equal weight to all the tiles in the captured image in identifying the location of the vehicle.
However, it does not consider the non-uniform noise levels in the sensed image induced by the geometry of the acquisition process.

Below, we describe a principled procedure to find the maximum a posteriori distribution of the location based on the captured image and the global map.
We shall see that this naturally leads to a weighted norm, which corresponds to the induced norm of a generalized inner product.

\begin{proposition}
Let the a priori probability of the true location be uniformly distributed on $[L]$.
Then, the maximum likelihood location estimate is given by
\begin{equation}
\hat{\ell} = \operatorname*{arg\;min}_{\ell\in[L]}
\sum_{k=1}^{N_w}\sum_{j=1}^{N_d} \frac{({y}_{k,j}^\ell-{y}_{k,j})^2}{2\sigma_i^2 + N_0/\tilde{A}_{k,j}} .
\end{equation}
\end{proposition}
\begin{IEEEproof}
Let $X$ be the random variable denoting the location of the vehicle, taking values in the set $[L]$.
Then, using the independence across tiles, the posterior probability for location $\ell \in [L]$ becomes
\begin{equation*}
\begin{split}
&\Pr (X=\ell \mid \mathbf{Y} = \mathbf{y},\mathbf{Y}^l = \mathbf{y}^l, l \in [L] ) \\
&= \frac{p(\mathbf{Y} = \mathbf{y} \mid X=\ell, \mathbf{Y}^l = \mathbf{y}^l, l \in [L]) p(X=\ell)}{p( \mathbf{Y} = \mathbf{y} \mid \mathbf{Y}^l = \mathbf{y}^l, l \in [L])} \\
&= \frac{p(\mathbf{Y} = \mathbf{y} \mid X=\ell, \mathbf{Y}^{\ell} = \mathbf{y}^{\ell}) p(X=\ell)}{p( \mathbf{Y} = \mathbf{y} \mid \mathbf{Y}^l = \mathbf{y}^l, l \in [L])} \\
&\propto p(\mathbf{Y} = \mathbf{y} \mid X=\ell, \mathbf{Y}^{\ell} = \mathbf{y}^{\ell}) \\
&= \prod_{k=1}^{N_w} \prod_{j=1}^{N_d} p(Y_{k,j} = y_{k,j} \mid X=\ell, Y^{\ell}_{k,j} = y^{\ell}_{k,j}) .
\end{split}
\end{equation*}
To proceed forward, we rely on the Gaussian model introduced above.
Specifically, we have
\begin{equation*}
\begin{split}
&p \left( Y_{k,j} = y_{k,j} \middle| X=\ell, Y^{\ell}_{k,j} = y^{\ell}_{k,j} \right) \\
&= p \left( a^\ell_{k,j} + n^i_{k,j} + n^s_{k,j} = y_{k,j} \middle| X=\ell, Y^{\ell}_{k,j} = {y}^{\ell}_{k,j} \right) \\
    &= \int_{\R} p \left( n^i_{k,j} + {n}^s_{k,j} = {y}_{k,j} - t, {a}^\ell_{k,j} = t \middle| Y^{\ell}_{k,j} = {y}^{\ell}_{k,j} \right) \d t \\
    & \propto \int_{\R} e^{-\frac{1}{2}\left(\frac{ \left( {y}_{k,j}-t \right)^2}{ \left( \sigma_{k,j}^i \right)^2 +  \left( \sigma_{k,j}^s \right)^2} +\frac{ \left( {t-{y}_{k,j}^\ell} \right)^2}{ \left( \sigma_{k,j}^i \right)^2}\right) } \d t \\
    & \propto \int_{\R} e^{-\frac{1}{2}\left(\frac{ \left( {{y}_{k,j}-{y}_{k,j}^\ell-x} \right)^2}{\sigma_i^2 + N_0/\tilde{A}_{k,j}} +\frac{{x}^2}{\sigma_i^2}\right) } \d x.
\end{split}
\end{equation*}
First, we observe that
\begin{equation*}
\begin{split}
\int_{t\in\R} e^{-\frac{1}{2}\left(at^2+bt + c \right)} \d t
&= e^{\frac{1}{8} \left( \frac{b^2}{a}-4c \right) } \int_{t\in\R} e^{-\frac{a}{2} \left( \left( t+\frac{b}{2a} \right)^2 \right) } \d t \\
&= \sqrt{\frac{2\pi}{a}} \exp \left( \frac{1}{8} \left( \frac{b^2}{a}-4c \right) \right).
\end{split}
\end{equation*}
This can be used to simplify the expression above.
We compare $ax^2+bx+c$ with the term inside the bracket in the last equation and write
\begin{align*}
ax^2 + bx + c
&= x^2 \left( \frac{1}{\sigma_i^2+ \big( \sigma_{k,j}^s \big)^2} + \frac{1}{\sigma_i^2} \right) \\
&\quad - x \frac{2 \big( {y}_{k,j}^\ell-{y}_{k,j} \big)}{\sigma_i^2+ \big( \sigma_{k,j}^s \big)^2}
+ \frac{\big( {y}_{k,j}^\ell-{y}_{k,j} \big)^2}{\sigma_i^2+ \big( \sigma_{k,j}^s \big)^2} .
\end{align*}
Therefore, we can compute
\begin{equation*}
\begin{split}
\frac{b^2}{a}-4c
&=- \frac{4 \left( {y}_{k,j}^\ell-{y}_{k,j} \right)^2}{2 \sigma_i^2+ \left( \sigma_{k,j}^s \right)^2} .
\end{split}
\end{equation*}
The likelihood becomes
\begin{equation*}
\begin{split}
&p \left( Y_{k,j} = y_{k,j} \middle| X=\ell, Y^{l}_{k,j} = y^{l}_{k,j}, l \in [L] \right) \\
&\quad \propto \exp \left( -\frac{1}{2} \cdot \frac{ \big( {y}_{k,j}^\ell-{y}_{k,j} \big)^2}{2 \sigma_i^2+ \big( \sigma_{k,j}^s  \big)^2} \right) .
\end{split}
\end{equation*}
Hence, we can compute the posterior for the location as
\begin{align*}
&p \left( X =\ell \middle| \mathbf{Y} = \mathbf{y}, \mathbf{Y}^{l} = \mathbf{y}^{l}, l \in [L] \right) \\
&\quad \propto \prod_{k=1}^{N_w} \prod_{j=1}^{N_d} \exp \left( -\frac{1}{2} \cdot \frac{\big({y}_{k,j}^\ell-{y}_{k,j}\big)^2}{2 \sigma_i^2+\big(\sigma_{k,j}^s\big)^2} \right) .
\end{align*}
Therefore, we can form the maximum a posteriori estimate of the location as
\begin{align*}
    \hat{\ell} &= \operatorname*{arg\;min}_{\ell \in [L]}
    \sum_{k=1}^{N_w} \sum_{j=1}^{N_d} \frac{\big( {y}_{k,j}^\ell-{y}_{k,j} \big)^2}{2 \sigma_i^2 + \big( \sigma_{k,j}^s \big)^2},
\end{align*}
where $(\sigma^s_{k,j})^2 = N_0/\tilde{A}_{k,j}$.
\end{IEEEproof}

One interpretation of this result stems from the vectorized observation $\vec{\mathbf{y}}$.
The maximum likelihood estimate of the true location can be obtained by minimizing a weighted inner product.
The actual weights of the inner product are determined by the projection of the road onto the focal plane of the camera or, equivalently, the non-uniform noise amplification that takes place during the perspective transformation of the acquired image.
We formalize this observation as a corollary below.

\begin{corollary}
Let the a priori probability of the true location be uniformly distributed on $[L]$.
Then, the maximum likelihood location estimate is given by minimizing the induced norm of a generalized inner product (GIP)
\begin{equation} \label{eq:EGIP}
\hat{\ell}_{\EGIP}
= \operatorname*{arg\;min}_{\ell\in[L]} \left\| \vec{\mathbf{y}}-\vec{\mathbf{y}}^\ell \right\|_{\mathbf{G}}
\end{equation}
where $\mathbf{G}$ is a diagonal matrix whose non-zero entries are determined by vectorization map \eqref{eq:vectorize}, which yields
\begin{equation}
\mathbf{G}_{k + (j-1) N_w, k + (j-1) N_w} = \frac{1}{2 \sigma_i^2 + N_0/\tilde{A}_{k,j}}
\end{equation}
for $k \in [N_w], j \in [N_d]$.
\end{corollary}

When the scene is noiseless or if the algorithm is agnostic to the fact that the scene may be noisy; then $\sigma_i^2 \rightarrow 0$ and the maximum likelihood detector simplifies slightly.

\begin{corollary}
Let the a priori probability of the true location be uniformly distributed on $[L]$, and assume $\sigma_i^2 = 0$.
Then, the maximum likelihood location estimate reduces to
\begin{equation} \label{eq:GIP}
\hat{\ell}_{\GIP}
= \operatorname*{arg\;min}_{\ell \in[L]} \left\| \bar{\mathbf{y}}-\bar{\mathbf{y}}^\ell \right\|_{\tilde{\mathbf{G}}}
\end{equation}
where $\tilde{\mathbf{G}}$ is a diagonal matrix whose non-zero entries are given by
\begin{equation}
\tilde{\mathbf{G}}_{k + (j-1) N_w, k + (j-1) N_w} = \frac{\tilde{A}_{k,j}}{N_0}
\end{equation}
for $k \in [N_w], j \in [N_d]$.
\end{corollary}

In either case, the selection criterion differs from the naive Euclidean norm found in \eqref{eq:EuclideanNorm}.
Again, this distinction stems from the fact that the focal plane of the camera and the surface of the road are misaligned, combined with the fact that the acquisition process is noisy.
The same approach can be extended to more intricate and practical scenarios, which is the aim of the next section.

\subsection{Enhanced Normalized Mutual Information}
\label{subsection:EnhancedNormalizedMutualInformation}

As mentioned in the introduction, the normalized mutual information (NMI) has been used for image registration and localization in the past.
In its original form, NMI does not account for the fact that the acquisition process and the global reference can be noisy.
Taking noise into consideration is especially important when the noise profile of the sensed image is not uniform across sections.
Much like the optimal solution above, we would like to derive an enhanced normalized mutual information (ENMI) performance criterion that accounts for noise in a principled fashion.
To the best of our knowledge, this aim is novel and it can lead to significant algorithmic improvements.

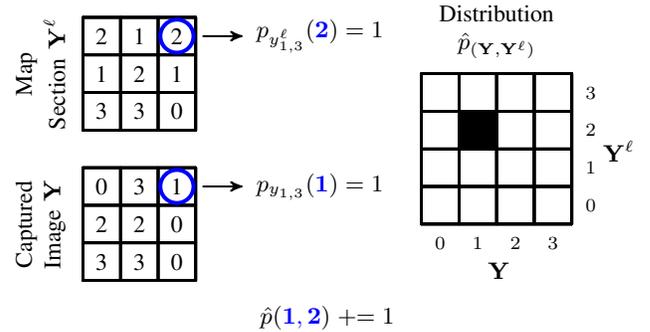
\begin{figure}[!ht]
\centering{\begin{tikzpicture}[
  font=\small, >=stealth', line width=1.0pt, line cap=round
]

\foreach \y in {0,1,2,3} {
  \foreach \x in {0,1,2,3} {
    \draw (0.5*\x, 0.25) -- (0.5*\x, 1.75);
    \draw (0, 0.25+0.5*\y) -- (1.5, 0.25+0.5*\y);
  }
}
\node[rotate=90] at (-0.75,1) {Captured};
\node[rotate=90] at (-0.375,1) {Image $\mathbf{Y}$};
\node (kvalue) at (1.25, 1.5) {1};
\node at (0.75, 1.5) {3};
\node at (0.25, 1.5) {0};
\node at (0.25, 1) {2};
\node at (0.75, 1) {2};
\node at (1.25, 1) {0};
\node at (0.25, 0.5) {3};
\node at (0.75, 0.5) {3};
\node at (1.25, 0.5) {0};

\foreach \y in {0,1,2,3} {
  \foreach \x in {0,1,2,3} {
    \draw (0.5*\x, 2.25) -- (0.5*\x, 3.75);
    \draw (0, 2.25+0.5*\y) -- (1.5, 2.25+0.5*\y);
  }
}
\node[rotate=90] at (-0.75,3) {Map};
\node[rotate=90] at (-0.375,3) {Section $\mathbf{Y}^\ell$};
\node (lvalue) at (1.25, 3.5) {2};
\node at (0.75, 3.5) {1};
\node at (0.25, 3.5) {2};
\node at (0.25, 3) {1};
\node at (0.75, 3) {2};
\node at (1.25, 3) {1};
\node at (0.25, 2.5) {3};
\node at (0.75, 2.5) {3};
\node at (1.25, 2.5) {0};

\node (pair1) at (3.15, 1.5) {$p_{y_{1,3}}(\mathbf{\textcolor{blue}{1}}) = 1$};

\draw[->, line width=0.75] (1.6,1.5) -- (2.125,1.5);

\node (pair2) at (3.15, 3.5) {$p_{y^{\ell}_{1,3}}(\mathbf{\textcolor{blue}{2}}) = 1$};

\draw[->, line width=0.75] (1.6,3.5) -- (2.125,3.5);

\draw[blue, line width=1.5] (1.25, 1.5) circle (0.22);
\draw[blue, line width=1.5] (1.25, 3.5) circle (0.22);

\foreach \y in {0,1,2,3,4} {
  \foreach \x in {0,1,2,3,4} {
    \draw (4.5+0.5*\x, 1.00) -- (4.5+0.5*\x, 3.00);
    \draw (4.5, 1+0.5*\y) -- (6.5, 1+0.5*\y);
  }
}

\node (value) at (3.25, -0.25) {$\hat{p}(\mathbf{\textcolor{blue}{1, 2}}) \mathrel{+}= 1$};

\node (J) at (5.5, 3.8) {Distribution};
\node (D) at (5.5, 3.375) {$\hat{p}_{(\mathbf{Y}, \mathbf{Y}^\ell)}$};

\foreach \p in {0,1,2,3} {
  \node at (6.75, 1.25+0.5*\p) {\scriptsize $\p$};
  \node at (4.75+0.5*\p, 0.75) {\scriptsize $\p$};
}
\draw (5.25, 0.375) node[anchor=west] {$\mathbf{Y}$};
\draw (7.125, 2.25) node[anchor=north] {$\mathbf{Y}^\ell$};


\draw[fill=black] (5, 2.5) rectangle (5.5, 2);




\end{tikzpicture}}
    \caption{This diagram shows how the probability mass is added to the empirical joint distribution in NMI.
    In the example above, the selected tiles yield $(1,2)$ and, hence, the corresponding location is incremented.
    Once all the tiles are accounted for, the matrix is normalized and the empirical joint distribution is obtained.
    For ease of exposition, we restricted the space of possibilities to $\mathcal{V} = \{0, 1, 2, 3\}$.}
    \label{fig:NMIIllustration}
\end{figure}

Before describing our proposed algorithm, we briefly review basic properties of NMI.
Computing NMI for two tiled images $\mathbf{Y}$ and $\mathbf{Y}^\ell$ is straightforward (see Fig.~\ref{fig:NMIIllustration}).
Given two tiled images, we pair the values corresponding to the tiles on both the images.
Then, we construct the joint empirical distribution on $\mathcal{V} \times \mathcal{V}$ by counting the instances of every possible pair of values, and then normalizing.
With the empirical joint distribution in hand, the NMI of the two images can be computed as
\begin{equation} \label{eq:NMI}
\NMI[\mathbf{Y}, \mathbf{Y}^\ell]
= \frac{\H [\mathbf{Y}] + \H [\mathbf{Y}^\ell]}{\H[\mathbf{Y}, \mathbf{Y}^\ell]},
\end{equation}
where $\H[\cdot]$ denotes the entropy of the empirical distribution associated with its argument.
An NMI-based localization algorithm computes the NMI of the captured image with every candidate section of the global map.
The section that has the highest NMI when combined with the captured image is declared to be the match.
The vehicle computes the location based on the meta data associated with this matched section.
We emphasize that, in the generation of the joint empirical distribution, the candidate images are quantized into a discrete set of values.

\begin{remark}
Although NMI has been used extensively in the literature, it may be useful to provide some insight about its underlying properties because some readers may be unfamiliar with the concept.
When the two arguments of NMI are the same tiled image, the support of the empirical distribution concentrates on the diagonal elements and the joint entropy remains small.
However, when two distinct images are used as arguments, the joint empirical distribution tends to be more diffused and the corresponding joint entropy is often much larger.
Similar concepts apply when the arguments are quantized versions of noisy images.
Interestingly, when a constant is added to all the tiles of an image, which corresponds loosely to a change in illumination, the diagonal support of the empirical distribution between the original image and the altered version shifts up or down; nevertheless, the empirical distribution remains concentrated and the joint entropy stays large.
This remark offers some motivation about the use of NMI in localization problems found in the literature.
\end{remark}

We now return to the task at hand, which is to account for noise while applying techniques similar to NMI.
Since the global map and the captured images are noisy, the signals in $(k,j)$th tiles of the acquired image and the global map section are distributed as
\begin{align}
{a}_{k,j} &\sim \mathcal{N} \left( {y}_{k,j}, (\sigma^{i}_{k,j})^2+(\sigma^{s}_{k,j})^2 \right) \\
{a}_{k,j}^\ell &\sim \mathcal{N} \left( {y}_{k,j}^\ell, (\sigma^{i}_{k,j})^2 \right)
\end{align}
where $\mathcal{N}(\mu, \sigma^2)$ denotes the Gaussian distribution with mean $\mu$ and variance $\sigma^2$.
We use the above distributions to propose the Enhanced Normalized Mutual Information (ENMI) criterion, which builds on the Normalized Mutual information (NMI) but accounts for noise.

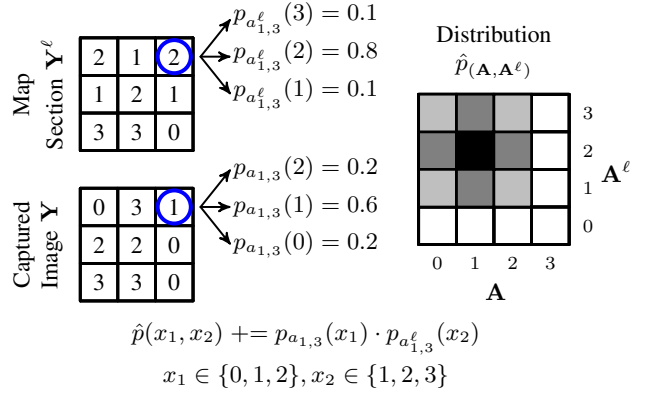
\begin{figure}[bt]
\centering{\begin{tikzpicture}[
  font=\small, >=stealth', line width=1.0pt, line cap=round
]

\foreach \y in {0,1,2,3} {
  \foreach \x in {0,1,2,3} {
    \draw (0.5*\x, 0.25) -- (0.5*\x, 1.75);
    \draw (0, 0.25+0.5*\y) -- (1.5, 0.25+0.5*\y);
  }
}
\node[rotate=90] at (-0.75,1) {Captured};
\node[rotate=90] at (-0.375,1) {Image $\mathbf{Y}$};
\node (kvalue) at (1.25, 1.5) {1};
\node at (0.75, 1.5) {3};
\node at (0.25, 1.5) {0};
\node at (0.25, 1) {2};
\node at (0.75, 1) {2};
\node at (1.25, 1) {0};
\node at (0.25, 0.5) {3};
\node at (0.75, 0.5) {3};
\node at (1.25, 0.5) {0};

\foreach \y in {0,1,2,3} {
  \foreach \x in {0,1,2,3} {
    \draw (0.5*\x, 2.25) -- (0.5*\x, 3.75);
    \draw (0, 2.25+0.5*\y) -- (1.5, 2.25+0.5*\y);
  }
}
\node[rotate=90] at (-0.75,3) {Map};
\node[rotate=90] at (-0.375,3) {Section $\mathbf{Y}^\ell$};
\node (lvalue) at (1.25, 3.5) {2};
\node at (0.75, 3.5) {1};
\node at (0.25, 3.5) {2};
\node at (0.25, 3) {1};
\node at (0.75, 3) {2};
\node at (1.25, 3) {1};
\node at (0.25, 2.5) {3};
\node at (0.75, 2.5) {3};
\node at (1.25, 2.5) {0};

\node (pair2) at (3, 2) {$p_{a_{1,3}}(2) = 0.2$};
\node (pair1) at (3, 1.5) {$p_{a_{1,3}}(1) = 0.6$};
\node (pair0) at (3, 1.0) {$p_{a_{1,3}}(0) = 0.2$};

\draw[->, line width=0.75] (1.6,1.5) -- (2,2.0);
\draw[->, line width=0.75] (1.6,1.5) -- (2,1.5);
\draw[->, line width=0.75] (1.6,1.5) -- (2,1.0);

\node (pair2) at (3, 4) {$p_{a^{\ell}_{1,3}}(3) = 0.1$};
\node (pair1) at (3, 3.5) {$p_{a^{\ell}_{1,3}}(2) = 0.8$};
\node (pair0) at (3, 3.0) {$p_{a^{\ell}_{1,3}}(1) = 0.1$};

\draw[->, line width=0.75] (1.6,3.5) -- (2,4);
\draw[->, line width=0.75] (1.6,3.5) -- (2,3.5);
\draw[->, line width=0.75] (1.6,3.5) -- (2,3.0);

\draw[blue, line width=1.5pt] (1.25, 1.5) circle (0.22);
\draw[blue, line width=1.5pt] (1.25, 3.5) circle (0.22);

\foreach \y in {0,1,2,3,4} {
  \foreach \x in {0,1,2,3,4} {
    \draw (4.5+0.5*\x, 1.00) -- (4.5+0.5*\x, 3.00);
    \draw (4.5, 1+0.5*\y) -- (6.5, 1+0.5*\y);
  }
}

\node (value) at (3, -0.25) {$\hat{p}(x_1, x_2) \mathrel{+}= p_{a_{1,3}}(x_1) \cdot p_{a^\ell_{1,3}}(x_2)$};
\node (value1) at (3, -0.75) {$x_1\in\{0,1,2\}, x_2\in\{1,2,3\}$};

\node (J) at (5.5, 3.8) {Distribution};
\node (D) at (5.5, 3.375) {$\hat{p}_{(\mathbf{A}, \mathbf{A}^\ell)}$};

\foreach \p in {0,1,2,3} {
  \node at (6.75, 1.25+0.5*\p) {\scriptsize $\p$};
  \node at (4.75+0.5*\p, 0.75) {\scriptsize $\p$};
}
\draw (5.25, 0.375) node[anchor=west] {$\mathbf{A}$};
\draw (7.125, 2.25) node[anchor=north] {$\mathbf{A}^\ell$};


\draw[fill=black] (5, 2.5) rectangle (5.5, 2);

\draw[fill=gray] (4.5, 2.5) rectangle (5, 2);
\draw[fill=gray] (5.5, 2.5) rectangle (6, 2);
\draw[fill=gray] (5.0, 2.5) rectangle (5.5, 3);
\draw[fill=gray] (5.0, 1.5) rectangle (5.5, 2);

\draw[fill=lightgray] (5.5, 3) rectangle (6, 2.5);
\draw[fill=lightgray] (5.5, 2) rectangle (6, 1.5);
\draw[fill=lightgray] (4.5, 3) rectangle (5, 2.5);
\draw[fill=lightgray] (4.5, 2) rectangle (5, 1.5);


\end{tikzpicture}}
    \caption{The diagram shows how the probability mass is added to the empirical joint distribution in the algorithm ENMI2D. Observe that the mass is distributed in the value-pair space in accordance with the uncertainty in the underlying signal in both the global map and the captured image.}
    \label{fig:ENMI2DIllustration}
\end{figure}

Since the images are noisy, we use the distribution of the underlying signal values to construct an empirical joint posterior probability distribution of the pair of values across the tiles.
This empirical distribution apportions the probability mass in the value-pair space according to the likelihood of the underlying signals in the pair of images.
Therefore, the mass corresponding to a single pair of tiles is spread in the value-pair space with the amount of spread depending on the variance of the noise (see Fig.~\ref{fig:ENMI2DIllustration}).
Using the ensuing posterior distribution, we can compute the 2D ENMI using the following equation
\begin{equation}
    \label{eq:ENMI2D}
    \DENMI[\mathbf{Y}, \mathbf{Y}^\ell] = \frac{\H[\mathbf{A}] + \H[\mathbf{A}^\ell]}{\H[\mathbf{A}, \mathbf{A}^\ell]}.
\end{equation}
Observe that in \eqref{eq:ENMI2D}, we are using the posterior distribution over the underlying signals rather than just the observed signals.

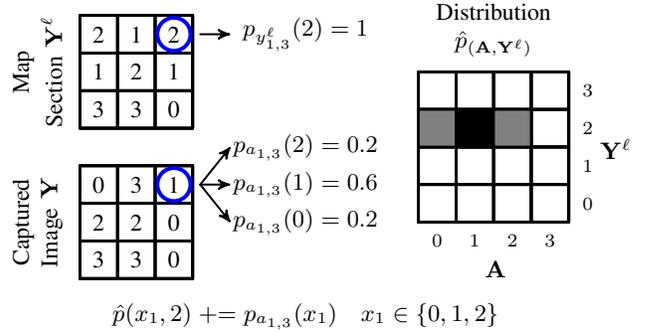
\begin{figure}[!tb]
\centering{\begin{tikzpicture}[
  font=\small, >=stealth', line width=1.0pt, line cap=round
]

\foreach \y in {0,1,2,3} {
  \foreach \x in {0,1,2,3} {
    \draw (0.5*\x, 0.25) -- (0.5*\x, 1.75);
    \draw (0, 0.25+0.5*\y) -- (1.5, 0.25+0.5*\y);
  }
}
\node[rotate=90] at (-0.75,1) {Captured};
\node[rotate=90] at (-0.375,1) {Image $\mathbf{Y}$};
\node (kvalue) at (1.25, 1.5) {1};
\node at (0.75, 1.5) {3};
\node at (0.25, 1.5) {0};
\node at (0.25, 1) {2};
\node at (0.75, 1) {2};
\node at (1.25, 1) {0};
\node at (0.25, 0.5) {3};
\node at (0.75, 0.5) {3};
\node at (1.25, 0.5) {0};

\foreach \y in {0,1,2,3} {
  \foreach \x in {0,1,2,3} {
    \draw (0.5*\x, 2.25) -- (0.5*\x, 3.75);
    \draw (0, 2.25+0.5*\y) -- (1.5, 2.25+0.5*\y);
  }
}
\node[rotate=90] at (-0.75,3) {Map};
\node[rotate=90] at (-0.375,3) {Section $\mathbf{Y}^\ell$};
\node (lvalue) at (1.25, 3.5) {2};
\node at (0.75, 3.5) {1};
\node at (0.25, 3.5) {2};
\node at (0.25, 3) {1};
\node at (0.75, 3) {2};
\node at (1.25, 3) {1};
\node at (0.25, 2.5) {3};
\node at (0.75, 2.5) {3};
\node at (1.25, 2.5) {0};

\node (pair2) at (3, 2) {$p_{a_{1,3}}(2) = 0.2$};
\node (pair1) at (3, 1.5) {$p_{a_{1,3}}(1) = 0.6$};
\node (pair0) at (3, 1.0) {$p_{a_{1,3}}(0) = 0.2$};

\draw[->, line width=0.75] (1.6,1.5) -- (2,2.0);
\draw[->, line width=0.75] (1.6,1.5) -- (2,1.5);
\draw[->, line width=0.75] (1.6,1.5) -- (2,1.0);

\node (pair1) at (3, 3.5) {$p_{y^{\ell}_{1,3}}(2) = 1$};
\draw[->, line width=0.75] (1.6,3.5) -- (2,3.5);

\draw[blue, line width=1.5pt] (1.25, 1.5) circle (0.22);
\draw[blue, line width=1.5pt] (1.25, 3.5) circle (0.22);

\foreach \y in {0,1,2,3,4} {
  \foreach \x in {0,1,2,3,4} {
    \draw (4.5+0.5*\x, 1.00) -- (4.5+0.5*\x, 3.00);
    \draw (4.5, 1+0.5*\y) -- (6.5, 1+0.5*\y);
  }
}

\node (value) at (3, -0.25) {$\hat{p}(x_1, 2) \mathrel{+}= p_{a_{1,3}}(x_1) \quad x_1\in\{0,1,2\}$};

\node (J) at (5.5, 3.8) {Distribution};
\node (D) at (5.5, 3.375) {$\hat{p}_{(\mathbf{A}, \mathbf{Y}^\ell)}$};

\foreach \p in {0,1,2,3} {
  \node at (6.75, 1.25+0.5*\p) {\scriptsize $\p$};
  \node at (4.75+0.5*\p, 0.75) {\scriptsize $\p$};
}
\draw (5.25, 0.375) node[anchor=west] {$\mathbf{A}$};
\draw (7.125, 2.25) node[anchor=north] {$\mathbf{Y}^\ell$};


\draw[fill=black] (5, 2.5) rectangle (5.5, 2);

\draw[fill=gray] (4.5, 2.5) rectangle (5, 2);
\draw[fill=gray] (5.5, 2.5) rectangle (6, 2);



\end{tikzpicture}}
    \caption{The diagram shows how the probability mass is added to the empirical joint distribution in the algorithm ENMI1D. Observe that the mass is distributed exclusively in the value space of the captured image, in accordance with the uncertainty in the underlying signal. This algorithm assumes that the global map is noiseless, and hence, full probability mass is placed on the observed value in the global map.}
    \label{fig:ENMI1DIllustration}
\end{figure}

If we assume that the global map is noiseless, perhaps due to statistical averaging over several acquisition rounds over a large period of time, then we get
\begin{equation*}
    \label{eq:ENMI}
    \ENMI[\mathbf{Y}, \mathbf{Y}^\ell] = \frac{\H[\mathbf{A}] + \H[\mathbf{Y}^\ell]}{\H[\mathbf{A}, \mathbf{Y}^\ell]}
\end{equation*}
where the allocation of the probability mass occurs only in the values of the captured image, hence the 1D nomenclature (see Fig.~\ref{fig:ENMI1DIllustration}).

Collectively, we refer to these decision-making algorithms, $\NMI$, $\ENMI$, $\DENMI$, as mutual information classifiers:
\begin{equation}
    \hat{\ell}_{{\rm MI}} = \operatorname*{arg\;max}_{\ell\in[L]} {\rm MI}[\mathbf{Y}, \mathbf{Y}^\ell] .
\end{equation}

\subsection{Probability of Misclassification}

To assess the performance of these alternatives, we are interested in the probability that a particular scheme makes an error under given circumstances.
In every case, the probability of misclassification error is given by
\begin{equation} \label{eq:ProbMisclassification}
    P_{e,\mathrm{A}} = \Pr \left( \hat{\ell}_{\mathrm{A}} \neq \ell \right)
\end{equation}
where $\ell$ is the true location.
The probability of misclassification in \eqref{eq:ProbMisclassification} may depend on noise variance and the number of candidate locations ($L$).
In building complete systems, these and other considerations must be taken into consideration.
Nevertheless, a goal of this article is to offer a fair comparative study of a potential algorithmic improvement, rather than build a full functional localization system.
As such, the abstract formulation above is sufficient to assess performance of the compared schemes in synthetic scenarios.
This is accomplished in the next section.

\section{Performance Evaluation and discussion}

The goal of this section is to assesss the value of the proposed approaches as algorithmic building blocks in larger systems.
Admittedly, it would be interesting to showcase the performance of the proposed changes in autonomous vehicles that already employ the Euclidean distance or NMI in their localization workflows.
However, integrating our findings in proprietary systems and then publishing results in an academic outlet is a logistical challenge that we are yet to solve.
As such, we focus the exposition on the intrinsic improvement associated with the proposed techniques in isolation, with the understanding that these building blocks can be integrated in larger workflows through software upgrades.
That is, the algorithmic opportunities presented in this article do not necessitate hardware alterations, rather they can be deployed through software updates.
Having said that, the proposed techniques could be used to optimize camera mount points in a principled manner, but this is beyond the scope of this article.
We illustrate the advantages of proposed algorithms through numerical simulations for various noise levels.
The parameters utilized throughout our numerical simulations are summarized in Table~\ref{table:NumericalParameters}.
In addition, monochromatic tile colors are quantized to eight bits, with $2^8 = 256$ shades and $\mathcal{V} = \{0, 1, \dots, 255\}$.
The mean tile value is set to $\mu = 128$; and the variance, to $\sigma = 5$.

\begin{table}[tbh]
\begin{center}
\begin{tabular}{||l|l|l||}
\hline
\textbf{Parameter} & \textbf{Value} & \textbf{Description} \tabularnewline \hline
Height & $h$ & 60~cm \tabularnewline
Depression angle & $\theta$ & $36^{\circ}$ \tabularnewline
Focal length & $f$ & 0.0367~cm \tabularnewline
Square tile sides & $s$ & 20~cm \tabularnewline \hline
Signal-to-intrinsic noise ratio & SINR & 3~dB (IP), 10~dB (MI) \tabularnewline
Signal-to-sensor noise ratio & $\sigma^2/N_0$ & 10--80, 45~dB (AR1) \tabularnewline \hline
Tile depth count & $N_d$ & 11 tiles \tabularnewline
Tile width count & $N_w$ & 6 tiles \tabularnewline \hline
\end{tabular}
\end{center}
\caption{These parameters are employed in the numerical simulations.}
\label{table:NumericalParameters}
\end{table}

We consider the elementary scenario where $L = 2$, and we look at the performance of the various classifiers $P_{e,\mathrm{A}}$ as a function of $N_0$.
Specifically, the random sections to be compared are created by independently generating every tile according to $a^\ell_{k,j} \sim \mathcal{N}(\mu, \sigma^2)$ and $n^i_{k,j} \sim \mathcal{N}(0, (\sigma_{k,j}^i)^2)$.
The effect of the perspective transformation is captured by adding noise to the sensed image, $n^s_{k,j} \sim \mathcal{N}(0, (\sigma_{k,j}^s)^2)$.
In both cases, we have $k \in [N_w]$ and $j \in [N_d]$.
This is consistent with the physical process described in Section~\ref{section:ImageAcquisition}.
Points in each of the figures are obtained by averaging 10,000 independent trials.

\subsection{Numerical Simulations for Maximum Likelihood}

\begin{figure}[tbh]
    \centering
    \begin{tikzpicture}

\begin{axis}[
font=\small,
scale only axis,
width=6.5cm,
height=4.75cm,
xmode=log,
ymin=-0.05, ymax=0.55,
xlabel={Noise Spectral Density $(N_0)$},
ylabel={Probability of Misclassification},
xmajorgrids,
ymajorgrids,
zmajorgrids,
legend entries={$\SIP$, $\GIP$, $\EGIP$},
legend style={nodes=right},
legend pos=north west]

\addplot [color=black, dotted, line width=1.5pt]
coordinates{
(2.5e-05, 0.0012)
(3.9622329811527855e-05, 0.0013)
(6.279716078773956e-05, 0.0036)
(9.952679263837424e-05, 0.0086)
(0.00015773933612004825, 0.0152)
(0.00025, 0.0311)
(0.0003962232981152785, 0.0675)
(0.0006279716078773956, 0.1137)
(0.0009952679263837423, 0.1846)
(0.0015773933612004826, 0.246)
(0.0025, 0.3112)
(0.003962232981152785, 0.3683)
(0.0062797160787739495, 0.3984)
(0.009952679263837432, 0.432)
(0.015773933612004826, 0.4578)
(0.025, 0.4804)
(0.03962232981152785, 0.4896)
(0.06279716078773949, 0.4889)
(0.09952679263837433, 0.5064)
(0.15773933612004823, 0.5021)
(0.25, 0.5075)
(0.39622329811527834, 0.4978)
(0.6279716078773949, 0.5027)
(0.9952679263837434, 0.5069)
(1.5773933612004833, 0.5083)
};
\addplot[color=black, densely dashed, line width=2.5pt]
coordinates{
(2.5e-05, 0.0789)
(3.9622329811527855e-05, 0.0769)
(6.279716078773956e-05, 0.0746)
(9.952679263837424e-05, 0.0785)
(0.00015773933612004825, 0.083)
(0.00025, 0.0923)
(0.0003962232981152785, 0.1015)
(0.0006279716078773956, 0.1117)
(0.0009952679263837423, 0.1336)
(0.0015773933612004826, 0.1707)
(0.0025, 0.2066)
(0.003962232981152785, 0.2597)
(0.0062797160787739495, 0.307)
(0.009952679263837432, 0.3606)
(0.015773933612004826, 0.4161)
(0.025, 0.4513)
(0.03962232981152785, 0.4839)
(0.06279716078773949, 0.486)
(0.09952679263837433, 0.4893)
(0.15773933612004823, 0.4981)
(0.25, 0.5013)
(0.39622329811527834, 0.4999)
(0.6279716078773949, 0.4932)
(0.9952679263837434, 0.5077)
(1.5773933612004833, 0.5006)
};
\addplot[color=black, line width=1.5pt]
coordinates{
(2.5e-05, 0.001)
(3.9622329811527855e-05, 0.0009)
(6.279716078773956e-05, 0.0029)
(9.952679263837424e-05, 0.0056)
(0.00015773933612004825, 0.0096)
(0.00025, 0.0146)
(0.0003962232981152785, 0.0302)
(0.0006279716078773956, 0.0518)
(0.0009952679263837423, 0.085)
(0.0015773933612004826, 0.133)
(0.0025, 0.1882)
(0.003962232981152785, 0.2436)
(0.0062797160787739495, 0.3007)
(0.009952679263837432, 0.358)
(0.015773933612004826, 0.4161)
(0.025, 0.4487)
(0.03962232981152785, 0.4845)
(0.06279716078773949, 0.4859)
(0.09952679263837433, 0.4902)
(0.15773933612004823, 0.498)
(0.25, 0.5016)
(0.39622329811527834, 0.5)
(0.6279716078773949, 0.4933)
(0.9952679263837434, 0.5078)
(1.5773933612004833, 0.5006)
};

\end{axis}
\end{tikzpicture}
    \caption{This plot compares the performance of the three inner-product-based (IP) algorithms.
    The algorithm that takes into account the unequal noise amplification due to the perspective transformation and the CCD noise outperforms the other two schemes, including the Euclidean distance which is used extensively in the literature.}
    \label{fig:ML}
\end{figure}
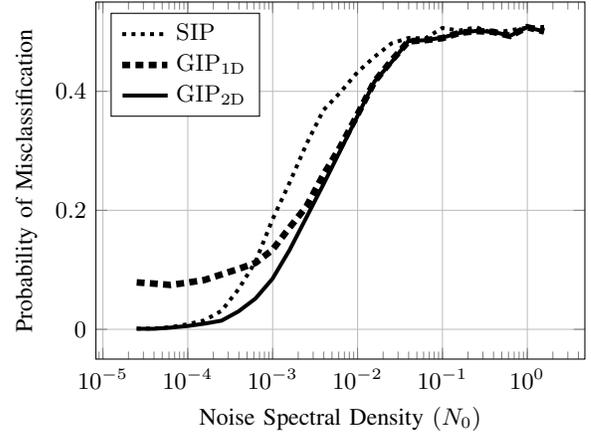

The first set of results we present pertain to the maximum likelihood detection scheme of Section~\ref{subsection:MaximumLikelihoodDetection}.
Therein, the selection criteria are based on the Euclidean distance and the distances induced by generalized inner products.
This includes the standard inner product of \eqref{eq:EuclideanNorm}, the 1D generalized product of \eqref{eq:GIP}, and the 2D generalized product of \eqref{eq:EGIP}.
Fig.~\ref{fig:ML} shows that the maximum likelihood algorithm that takes advantage of unequal noise amplification and accounts for sensing noise, $\EGIP$, uniformly outperforms the two alternatives for all values of the noise power spectral density $N_0$.

This figure also illustrates that disregarding the randomness associated with rapidly changing visual attributes of the road surface, $\sigma_i^2$, can lead to a brittle solutions.
This is manifested in the graph through the poor performance of $\GIP$ under some conditions.
The Euclidean distance is more robust, nevertheless it offers worse performance compared to the $\EGIP$ algorithm.

\begin{figure}[tbh]
    \centering
    \begin{tikzpicture}

\begin{axis}[
font=\small,
scale only axis,
width=6.5cm,
height=4.75cm,
xmin=-0.05, xmax=1.05,
ymin=-0.05, ymax=0.55,
xlabel={Auto correlation coefficient $\alpha$},
ylabel={Probability of Misclassification},
xmajorgrids,
ymajorgrids,
zmajorgrids,
legend entries={$\SIP$, $\GIP$, $\EGIP$},
legend style={nodes=right},
legend pos=north west]

\addplot [color=black, dotted, line width=1.5pt]
coordinates{
(0.05, 0.1459)
(0.1, 0.1429)
(0.15000000000000002, 0.1427)
(0.2, 0.15)
(0.25, 0.152)
(0.3, 0.1483)
(0.35000000000000003, 0.147)
(0.4, 0.1546)
(0.45, 0.1494)
(0.5, 0.1467)
(0.55, 0.1565)
(0.6000000000000001, 0.1471)
(0.6500000000000001, 0.1595)
(0.7000000000000001, 0.1548)
(0.7500000000000001, 0.1599)
(0.8, 0.169)
(0.8500000000000001, 0.1836)
(0.9000000000000001, 0.2154)
(0.95, 0.2822)
(0.96, 0.3176)
(0.97, 0.3464)
(0.98, 0.3922)
(0.99, 0.4286)
};
\addplot[color=black, densely dashed, line width=2.5pt]
coordinates{
(0.05, 0.1259)
(0.1, 0.118)
(0.15000000000000002, 0.1152)
(0.2, 0.1248)
(0.25, 0.1213)
(0.3, 0.1263)
(0.35000000000000003, 0.124)
(0.4, 0.1228)
(0.45, 0.1236)
(0.5, 0.1233)
(0.55, 0.122)
(0.6000000000000001, 0.1292)
(0.6500000000000001, 0.1208)
(0.7000000000000001, 0.1241)
(0.7500000000000001, 0.127)
(0.8, 0.1341)
(0.8500000000000001, 0.1501)
(0.9000000000000001, 0.1798)
(0.95, 0.2321)
(0.96, 0.2625)
(0.97, 0.2952)
(0.98, 0.3362)
(0.99, 0.403)
};
\addplot[color=black, line width=1.5pt]
coordinates{
(0.05, 0.0674)
(0.1, 0.0656)
(0.15000000000000002, 0.0673)
(0.2, 0.0696)
(0.25, 0.0701)
(0.3, 0.0678)
(0.35000000000000003, 0.0679)
(0.4, 0.0717)
(0.45, 0.0722)
(0.5, 0.0669)
(0.55, 0.0693)
(0.6000000000000001, 0.0704)
(0.6500000000000001, 0.0755)
(0.7000000000000001, 0.0761)
(0.7500000000000001, 0.0811)
(0.8, 0.0856)
(0.8500000000000001, 0.1014)
(0.9000000000000001, 0.1249)
(0.95, 0.1913)
(0.96, 0.224)
(0.97, 0.2587)
(0.98, 0.3194)
(0.99, 0.3881)
};

\end{axis}
\end{tikzpicture}
    \caption{Roads surfaces often have features that are spatially correlated.
    This plot shows that correlation degrades performance, however it does not affect the suitability of the three schemes.
    Furthermore, $\EGIP$ outperforms the other algorithms irrespective of the correlation coefficient.}
    \label{fig:AR1_ML}
\end{figure}
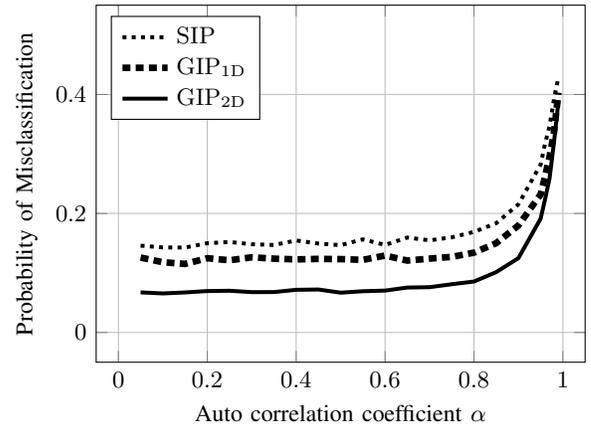
In nature, roads tend to have features that vary slowly.
Consequently, it is important to demonstrate that the proposed algorithm are robust to correlation in the color road surface.
To do so, we consider a scenario where the color of tiles along the depth of the road surface are generated by a stationary auto-regressive process, AR-1, whose governing parameter is $\alpha$.
That is, the road surface is created by a quantized version of the AR-1 equation
\begin{equation} \label{eq:AR1}
a_{k, j} = \alpha a_{k, j-1} + \sqrt{1 - \alpha^2} z_{k, j} \quad j \in [N_d - 1]
\end{equation}
where $z_{k, j} \sim \mathcal{N} (\mu, \sigma_a^2)$.
The initialization of the process must also be done carefully, as to ensure that the process is stationary.
This is standard for AR-1 processes.
Fig.~\ref{fig:AR1_ML} reveals that, although correlation affects performance, it does not undermine the suitability of the inner-product-based (IP) algorithms.
Furthermore, $\EGIP$ remains the best performer irrespective of $\alpha$, which is highly desirable.

\subsection{Numerical Simulations for Mutual Information}

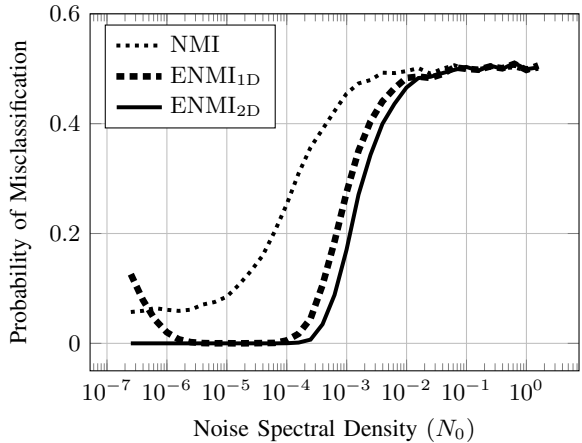
\begin{figure}[tbh]
    \centering
    \begin{tikzpicture}

\begin{axis}[
font=\small,
scale only axis,
width=6.5cm,
height=4.75cm,
xmode=log,
ymin=-0.05, ymax=0.6,
xlabel={Noise Spectral Density $(N_0)$},
ylabel={Probability of Misclassification},
xmajorgrids,
ymajorgrids,
zmajorgrids,
legend entries={$\NMI$, $\ENMI$, $\DENMI$},
legend style={nodes=right},
legend pos=north west]

\addplot [color=black, dotted, line width=1.5pt]
coordinates{
(2.5e-07, 0.0571)
(3.9622329811527857e-07, 0.0592)
(6.279716078773956e-07, 0.0642)
(9.952679263837423e-07, 0.0611)
(1.5773933612004824e-06, 0.0591)
(2.5e-06, 0.0613)
(3.962232981152785e-06, 0.0706)
(6.2797160787739555e-06, 0.0753)
(9.952679263837422e-06, 0.0848)
(1.5773933612004823e-05, 0.1062)
(2.5e-05, 0.1311)
(3.9622329811527855e-05, 0.1598)
(6.279716078773956e-05, 0.2035)
(9.952679263837424e-05, 0.2527)
(0.00015773933612004825, 0.3094)
(0.00025, 0.3568)
(0.0003962232981152785, 0.3899)
(0.0006279716078773956, 0.4229)
(0.0009952679263837423, 0.4547)
(0.0015773933612004826, 0.4732)
(0.0025, 0.4795)
(0.003962232981152785, 0.4928)
(0.0062797160787739495, 0.4921)
(0.009952679263837432, 0.4957)
(0.015773933612004826, 0.5011)
(0.025, 0.491)
(0.03962232981152785, 0.5)
(0.06279716078773949, 0.5063)
(0.09952679263837433, 0.4983)
(0.15773933612004823, 0.4995)
(0.25, 0.5032)
(0.39622329811527834, 0.501)
(0.6279716078773949, 0.5024)
(0.9952679263837434, 0.4966)
(1.5773933612004833, 0.4998)
};
\addplot[color=black, densely dashed, line width=2.5pt]
coordinates{
(2.5e-07, 0.1259)
(3.9622329811527857e-07, 0.0793)
(6.279716078773956e-07, 0.0435)
(9.952679263837423e-07, 0.0202)
(1.5773933612004824e-06, 0.0073)
(2.5e-06, 0.0022)
(3.962232981152785e-06, 0.0005)
(6.2797160787739555e-06, 0.0001)
(9.952679263837422e-06, 0.0002)
(1.5773933612004823e-05, 0.0)
(2.5e-05, 0.0002)
(3.9622329811527855e-05, 0.0005)
(6.279716078773956e-05, 0.0008)
(9.952679263837424e-05, 0.0052)
(0.00015773933612004825, 0.0174)
(0.00025, 0.0451)
(0.0003962232981152785, 0.1063)
(0.0006279716078773956, 0.1855)
(0.0009952679263837423, 0.2746)
(0.0015773933612004826, 0.3505)
(0.0025, 0.4042)
(0.003962232981152785, 0.4406)
(0.0062797160787739495, 0.4645)
(0.009952679263837432, 0.4833)
(0.015773933612004826, 0.4856)
(0.025, 0.4833)
(0.03962232981152785, 0.4903)
(0.06279716078773949, 0.5025)
(0.09952679263837433, 0.4998)
(0.15773933612004823, 0.4982)
(0.25, 0.5042)
(0.39622329811527834, 0.5008)
(0.6279716078773949, 0.5088)
(0.9952679263837434, 0.4979)
(1.5773933612004833, 0.506)
};
\addplot[color=black, line width=1.5pt]
coordinates{
(2.5e-07, 0.0)
(3.9622329811527857e-07, 0.0)
(6.279716078773956e-07, 0.0)
(9.952679263837423e-07, 0.0)
(1.5773933612004824e-06, 0.0)
(2.5e-06, 0.0)
(3.962232981152785e-06, 0.0)
(6.2797160787739555e-06, 0.0)
(9.952679263837422e-06, 0.0)
(1.5773933612004823e-05, 0.0)
(2.5e-05, 0.0)
(3.9622329811527855e-05, 0.0)
(6.279716078773956e-05, 0.0)
(9.952679263837424e-05, 0.0002)
(0.00015773933612004825, 0.0014)
(0.00025, 0.0068)
(0.0003962232981152785, 0.0347)
(0.0006279716078773956, 0.0882)
(0.0009952679263837423, 0.1703)
(0.0015773933612004826, 0.2708)
(0.0025, 0.3432)
(0.003962232981152785, 0.3998)
(0.0062797160787739495, 0.4361)
(0.009952679263837432, 0.4657)
(0.015773933612004826, 0.4831)
(0.025, 0.4853)
(0.03962232981152785, 0.492)
(0.06279716078773949, 0.4979)
(0.09952679263837433, 0.5033)
(0.15773933612004823, 0.4964)
(0.25, 0.5058)
(0.39622329811527834, 0.4996)
(0.6279716078773949, 0.5104)
(0.9952679263837434, 0.4991)
(1.5773933612004833, 0.5025)
};

\end{axis}
\end{tikzpicture}
    \caption{The two enhance NMI schemes take advantage of the unequal noise amplification that take place during a perspective transformation and, as such, they outperform the class NMI algorithm.
    Furthermore, $\DENMI$ accounts for sensing noise and performs uniformly better than the two alternatives, regardless of the sensing noise level.}
    \label{fig:ENMI}
\end{figure}

As mentioned above, NMI can be found as a building block in many localization applications because of its resilience to variations in illumination.
In Section~\ref{subsection:EnhancedNormalizedMutualInformation}, we proposed two new algorithms that better take into consideration the unequal noise amplification that takes place when images are rectified.
Conceptually, these two alternate algorithms build on NMI, but they are computationally more sophisticated and hence more demanding.
Still, they offer significant performance gains.
Fig.~\ref{fig:ENMI} plots probability of misclassification for the three mutual-information-based algorithms, $\NMI$, $\ENMI$, and $\DENMI$, as functions of the noise power spectral density $N_0$.
The latter two scheme take advantage of unequal noise amplification, and $\DENMI$ accounts for sensing noise. These two enhanced algorithms uniformly outperform $\NMI$, with $\DENMI$ showcasing the best performance.

Again, we offer a brief study of performance degradation related to correlation in the road surface, which translates into a correlated $\mathbf{A}^\ell$ in our simulation framework.
We examine the scenario described in \eqref{eq:AR1} and look for robustness.
The numerical simulations for mutual-information-based schemes show that correlation increases the probability of misclassification.
We note that the entropy of the randomly generated images decreases as a function of the auto-regressive parameter $\alpha$ and, as such, it becomes statistically more difficult to discriminate the true image from its counterpart.
This explains why the functions increase noticeably as $\alpha$ approaches one.
More importantly, performance varies smoothly as a function of $\alpha$, and $\DENMI$ remains uniformly better compares to the alternatives.
That is, the proposed algorithmic improvement based on signal processing techniques outperform the more naive schemes in all conditions.
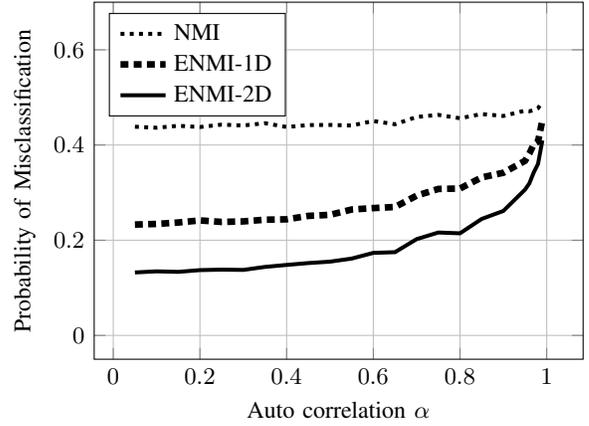
\begin{figure}[tbh]
    \centering
    \begin{tikzpicture}

\begin{axis}[
font=\small,
scale only axis,
width=6.5cm,
height=4.75cm,
ymin=-0.05, ymax=0.7,
xlabel={Auto correlation $\alpha$},
ylabel={Probability of Misclassification},
xmajorgrids,
ymajorgrids,
zmajorgrids,
legend entries={NMI, ENMI-1D, ENMI-2D},
legend style={nodes=right},
legend pos=north west]

\addplot [color=black, dotted, line width=1.5pt]
coordinates{
(0.05, 0.4384)
(0.1, 0.4363)
(0.15000000000000002, 0.4404)
(0.2, 0.4378)
(0.25, 0.4428)
(0.3, 0.4411)
(0.35000000000000003, 0.4458)
(0.4, 0.4379)
(0.45, 0.4419)
(0.5, 0.4422)
(0.55, 0.441)
(0.6000000000000001, 0.4505)
(0.6500000000000001, 0.4436)
(0.7000000000000001, 0.459)
(0.7500000000000001, 0.4633)
(0.8, 0.4562)
(0.8500000000000001, 0.4651)
(0.9000000000000001, 0.461)
(0.95, 0.4723)
(0.96, 0.471)
(0.97, 0.4731)
(0.98, 0.4773)
(0.99, 0.4909)
};
\addplot[color=black, densely dashed, line width=2.5pt]
coordinates{
(0.05, 0.2329)
(0.1, 0.2342)
(0.15000000000000002, 0.2372)
(0.2, 0.2416)
(0.25, 0.2384)
(0.3, 0.2395)
(0.35000000000000003, 0.2431)
(0.4, 0.2439)
(0.45, 0.2511)
(0.5, 0.2533)
(0.55, 0.2645)
(0.6000000000000001, 0.2675)
(0.6500000000000001, 0.2698)
(0.7000000000000001, 0.2937)
(0.7500000000000001, 0.3077)
(0.8, 0.3086)
(0.8500000000000001, 0.3319)
(0.9000000000000001, 0.3417)
(0.95, 0.3668)
(0.96, 0.382)
(0.97, 0.4012)
(0.98, 0.4072)
(0.99, 0.4519)
};
\addplot[color=black, line width=1.5pt]
coordinates{
(0.05, 0.1323)
(0.1, 0.1346)
(0.15000000000000002, 0.1335)
(0.2, 0.1372)
(0.25, 0.1382)
(0.3, 0.1376)
(0.35000000000000003, 0.1439)
(0.4, 0.1481)
(0.45, 0.152)
(0.5, 0.155)
(0.55, 0.1614)
(0.6000000000000001, 0.1734)
(0.6500000000000001, 0.1749)
(0.7000000000000001, 0.2023)
(0.7500000000000001, 0.2162)
(0.8, 0.2145)
(0.8500000000000001, 0.2447)
(0.9000000000000001, 0.2615)
(0.95, 0.3063)
(0.96, 0.3194)
(0.97, 0.3421)
(0.98, 0.3602)
(0.99, 0.4096)
};

\end{axis}
\end{tikzpicture}
    \caption{This graph tracks the performance of the three mutual-information-based (MI) algorithms as a function of correlation in the road surface.
    Our findings show that the proposed algorithmic enhancements are suitable for all correlation levels.
    Furthermore, $\DENMI$ uniformly outperforms alternatives.}
    \label{fig:AR1_ENMI}
\end{figure}

\section{Conclusion}

This article explores the implications of perspective transformations inherent to certain localization tasks, highlighting how these transformations can lead to uneven noise amplification.
We propose algorithmic methods to address this issue, resulting in performance enhancements that do not require expensive hardware modifications.
The value of this approach is demonstrated using simple classification tasks, showcasing the potential for integration into more complex and proprietary localization systems.
The foundational approach used in the articles can potentially be applied to multi-camera systems and LiDAR, although the treatment of these alternate scenarios is beyond the scope of the paper.

A key contribution of the article is the extension of normalized mutual information (NMI) to handle noisy environments effectively.
The proposed approach, termed enhanced normalized mutual information (ENMI), leverages Bayesian probability to address scenarios where images have unequal signal-to-noise ratios.
ENMI overcomes the limitations associated with quantization and pixelization, reframing image pre-processing as a natural tradeoff between performance and computational complexity.
Importantly, the technique preserves the desirable properties of NMI, including robustness and shift invariance, making it a powerful tool for applications in noisy and uneven imaging conditions.

The article offers several promising directions for future work.
These include developing a principled approach to optimize camera placement on autonomous vehicles, assessing the potential benefits of deploying multiple cameras, and extending the ENMI framework to handle LiDAR point clouds.
Another intriguing avenue is exploring how ENMI can be adapted for sequences of images or video segments, potentially enhancing its utility in dynamic scenarios.
Additionally, revisiting other applications of NMI, such as image registration, may reveal opportunities to improve performance by accounting for the impact of noisy sensors in these contexts, further broadening the scope of this approach.

\bibliographystyle{IEEEtran}
\bibliography{journalPaper,EnhancedNMI}

\end{document}